# Fast Learning From Sparse Data

**David Maxwell Chickering and David Heckerman Microsoft Research**
Redmond WA 98052-6399
dmax,heckerma@microsoft.com

## Abstract

We describe two techniques that significantly improve the running time of several standard machine-learning algorithms when data is sparse. The first technique is an algorithm that efficiently extracts one-way and two-way counts—either real or expected—from discrete data. Extracting such counts is a fundamental step in learning algorithms for constructing a variety of models including decision trees, decision graphs, Bayesian networks, and naive-Bayes clustering models. The second technique is an algorithm that efficiently performs the E-step of the EM algorithm (i.e., inference) when applied to a naive-Bayes clustering model. Using real-world data sets, we demonstrate a dramatic decrease in running time for algorithms that incorporate these techniques.

## 1 INTRODUCTION

Many real-world data sets represent domains for which there are hundreds or thousands of variables of interest. Furthermore, these data sets can contain thousands or millions of records. Statistical analyses of such data sets typically require all (or a significant sample) of the records to be resident in the main memory of a computer. Even with the growing memory sizes of modern-day machines, a large domain can be problematic if each record needs to explicitly store a value for every variable in the domain. Fortunately, for many large data sets, many or all of the variables appear in the same state for a large fraction of the records.

Consider a study of television-watching behavior for a population of viewers. For simplicity, assume that the variables of interest are binary indicators of whether or not a person watched each show during some time period (e.g. a week). Such a study would probably include hundreds of television shows and thousands of viewers. Each record in the data set encodes either "watched" or "didn't watch" for each one of the hundreds of shows. Any particular viewer, however, probably only watched a handful of television shows, and consequently the majority of the values for the corresponding record will be "didn't watch".

It seems obvious in this example that a *dense* representation of the data—where each record explicitly contains a value for every television show—is extremely inefficient. A better approach is to store, for each viewer, a (relatively short) list of shows that they watched. Such a *sparse* representation implicitly encodes the same information as the dense representation.

Regardless of the representation of the data, many standard machine-learning algorithms take a dense *view* of the data set. That is, these algorithms explicitly "request" values of variables in the records without regard to the representation of the data. It is straightforward to implement such a view for a sparse data set: whenever an algorithm requests the value for a variable that is not explicitly stored in a record, the default value ("didn't watch" in our example) is returned. The problem with hiding the sparse representation this way is we lose the ability to exploit the fact that the data is sparse in the implementation of the algorithms.

In this paper, we describe two techniques that allow several standard machine-learning algorithms to take advantage of sparse data. These techniques apply to algorithms that use models whose sufficient statistics—either real or expected—are one-way and two-way counts. Examples include (1) decision-tree learning algorithms (e.g. Breiman, 1984), (2) decision-graph learning algorithms (e.g. Chickering et al., 1997), (3) structure initialization for Bayesian-network learning algorithms (e.g. Cooper and Herskovitz, 1991, Buntine, 1991, Spiegelhalter et al., 1993, Heckerman et al., 1995), and (4) model-based clustering algorithms (e.g. Cheeseman and Stutz, 1995). The second technique we describe is a method to speed up inference in the E-step of the EM algorithm. We show that when used in conjunction with the first technique, we can speed up clustering algorithms dramatically.

Our method for efficiently extracting one-way and two-way counts from sparse data uses a trick that Moore and Lee (1998) use for caching all possible $n$-way counts in memory using an *AD*tree data structure. The algorithms we consider in this paper could, in fact, be applied to a data set that is represented using an *AD*tree. The emphasis of this paper, however, is modifying existing algorithms to take advantage of sparse data, and not on the representation of the data itself. In addition, our modifications allow efficient extraction of *expected* one-way and two-way counts.

In Section 2, we present notation, definitions, and background material. In Section 3, we describe how to efficiently extract one-way and two-way counts from sparse data. In Section 4, we show how to take advantage of sparse data when performing inference in the E-step of the EM algorithm. In Section 5, we provide experimental results that demonstrate a significant reduction in running time for machine-learning algorithms applied to sparse data that use the techniques described in this paper.

## 2 BACKGROUND AND NOTATION

In this paper, we assume that all variables in the domain of interest are discrete. For many applications, this assumption is not tremendously restrictive, as continuous variables can be discretized dynamically and the techniques we discuss apply directly.

We use upper case letters (e.g. $X_i$) to denote variables, and lower case letters (e.g. $x_i$) to denote values of variables. We use bold-face letters (e.g. **X**) to denote sets of variables. We use $r_i$ to denote the number of states for variable $X_i$.

There are many ways to represent sparse data efficiently, both in memory and on disk. The results of this paper apply to any representation with the following two properties:

- Each variable $X_i$ in the domain has a corresponding *default value* $d_i$. Whenever a variable occurs in its default value in a particular record, the value of that variable is not explicitly stored in the representation.

- For any record in the datatset, we can extract the values for the set of all variables that do not occur with their default value in time that is proportional to the number of variables in that set.

The first condition ensures that the size of the representation is proportional to the number of non-default values in the data set. The second condition ensures that we can efficiently extract the non-default values.

To simplify presentation, we assume for the remainder of the paper that each record in the data set is represented as a linked-list of variable/value pairs $(X_i, x_i)$, where $x_i \neq d_i$. This simple representation clearly satisfies the requirements above.

As mentioned in the introduction, we will describe a method to extract *one-way* and *two-way* counts efficiently. The one-way counts for a variable is a set that stores the number of times the variable occurs (marginally) in each of its states in the data. We use **SS**$(X_i)$ to denote the set of one-way counts for variable $X_i$, and we use $SS(X_i = x_i)$ to denote the number of cases in which $X_i = x_i$ in the data set. Two-way counts for a pair of discrete variables $X_i$ and $X_j$ encode, for each unique combination of $(x_i, x_j)$, the number of records in the data set for which $X_i = x_i$ and $X_j = x_j$. We use **SS**$(X_i, X_j)$ to denote the set of two-way counts for $X_i$ and $X_j$. To denote a specific count in this set, we use $SS(X_i = x_i, X_j = x_j)$.

| Record | A | B | C | | Record | List |
|--------|---|---|---|---|--------|------|
| 1 | 1 | 1 | 0 | | 1 | $\{(B,1)\}$ |
| 2 | 2 | 2 | 0 | | 2 | $\{(A,2),(B,2)\}$ |
| 3 | 1 | 0 | 2 | | 3 | $\{(C,2)\}$ |
| 4 | 1 | 2 | 1 | | 4 | $\{(B,2),(C,1)\}$ |
| 5 | 0 | 0 | 0 | | 5 | $\{(A,0)\}$ |
| 6 | 1 | 0 | 2 | | 6 | $\{(C,2)\}$ |
| 7 | 1 | 0 | 0 | | 7 | $\{\}$ |
| (a) | | | | | (b) | |

Figure 1: Two representations for a data set: (a) dense representation and (b) corresponding linked list. Default values for $A$, $B$, and $C$ are 1, 0 and 0, respectively.

Consider the matrix shown in Figure 1a, which represents a *dense view* of a small data set defined over three variables $A$, $B$, and $C$, where all variables have three states. To store the given data set in the sparse representation most efficiently, we define the default value for each variable to be the value that occurs most frequently in the data. Thus, the default value for $A$, $B$ and $C$ in the example would be 1, 0 and 0 respectively. The resulting linked-list sparse representation for the data is shown in Figure 1b. Figure 2a shows the two-way counts **SS**$(A, C)$, and Figure 2b shows the one-way counts **SS**$(C)$.

|       | $A=0$ | $A=1$ | $A=2$ |
|-------|-------|-------|-------|
| $C=0$ | 1     | 2     | 1     |
| $C=1$ | 0     | 1     | 0     |
| $C=2$ | 0     | 2     | 0     |

(a)

|       |   |
|-------|---|
| $C=0$ | 4 |
| $C=1$ | 1 |
| $C=2$ | 2 |

(b)

Figure 2: (a) Two-way counts $\mathbf{SS}(A,C)$ and (b) one-way counts $\mathbf{SS}(C)$ for the data set shown in Figure 1

## 3 EFFICIENTLY EXTRACTING TWO-WAY COUNTS

In this section, we describe a method to extract one-way and two-way counts from sparse data efficiently. We concentrate on the task of (1) extracting the one-way counts for all variables, and (2) extracting the two-way counts between a single *target* variable $T$ and *all other* variables $\mathbf{X} = \{X_1, ..., X_n\}$ in the domain. These two operations are fundamental to decision-tree learning algorithms when evaluating potential splits on a leaf node. A straightforward extension to the method yields the two-way counts between every pair of variables in the domain, which can be used to identify a maximum-branching (Edmonds, 1967) Bayesian network. As we shall see in Section 4, our method is also relevant to naive-Bayes clustering.

We use $t$ to denote a state of $T$, and $t_d$ to denote the default state of $T$.

The basic idea behind the method is to accumulate the counts only for variables that *do not* occur in their corresponding default states. We then *derive* the remaining counts using the known number of cases $m$. This trick was used by Moore and Lee (1998) to reduce the size of a data-structure they use to store all $n$-way counts from a data set.

A simple (dense-view) algorithm to extract the two-way counts

$$\{\mathbf{SS}(T,X_1), ..., \mathbf{SS}(T,X_n)\}$$

can be described as follows. First we initialize, for each value $i$, all entries in $\mathbf{SS}(T, X_i)$ to zero. Then, for each record in the data set, we add one to the relevant element in every set $\mathbf{SS}(T, X_i)$. That is, if $T = t$ and $X_i = x_i$ in the record, we increment $SS(T = t, X_i = x_i)$ by one. Note that we can easily calculate the one-way counts as well during the scan of the data set.

Let $m$ denote the number of records in the data set. Then the time to calculate the one-way and two-way counts, as a function of $m$ and $n$, is proportional to $m \cdot n$. Note that $m \cdot n$ is the total number of values in a dense representation of the data.

Recall that, in the sparse representation, we only explicitly store a value for a variable in a record if that value is not the default value for the variable. Let $l$ denote the total number of (non-default) values in the sparse representation. Figure 3 shows our algorithm that can extract all one-way and two-way counts in time proportional to $l$. In the next section, we show how the algorithm can also speed up the extraction of *expected* sufficient statistics.

For large data sets, the time to complete the algorithm in Figure 3 is dominated by the scan of the data in Step 2. Consequently, if only $l$ out of the $m \cdot n$ possible values in the data are non-default values, the optimization yields an algorithm that is roughly $\frac{m \cdot n}{l}$ times faster than the simple algorithm that checks the value for every variable in every record. In Section 5, we verify that speedups of this magnitude are achieved on real-world data sets.

## 4 CLUSTERING

In this section, we describe a method to efficiently cluster a sparse data set. In particular, we show how to perform inference in the E-step of the EM algorithm (Dempster, Laird and Rubin, 1977) such that the step completes in time that is proportional to the number of values explicitly stored in the sparse representation of the data set. We concentrate on the EM algorithm applied to a naive-Bayes model, shown in Figure 4, although the method applies to other clustering algorithms (such as K-means) and to other models whose expected sufficient statistics are one-way and two-way counts. The naive-Bayes model has been used by Cheeseman and Stutz (1995) in their system called AutoClass, and has been studied in depth by statisticians (e.g. Clogg, 1995).

The EM algorithm is a method that can be used to determine the maximum a posteriori or maximum likelihood configuration for the parameter values contained in a naive-Bayes model. Once identified, we can use these parameters to (fractionally) assign cluster membership to each record in the data. The resulting parameter values can also be used to approximate the marginal likelihood of the model, which allows learning algorithms to automatically identify the number of states for the cluster variable (Chickering and Heckerman, 1997).

To describe the EM algorithm, we adopt the following notation. We use $\mathbf{\Theta}$ to denote the set of all parameters stored in the model, and use $r_C$ to denote the number of states of the cluster node (i.e. the number of clusters for the model). The cluster node stores parameters $\{\theta_{C=1}, ..., \theta_{C=r_C}\}$, specifying the prior probability of a record (or person) belonging to any particular cluster.

1. Initialize to zero all counts in the sets
   $\{\mathbf{SS}(T), \mathbf{SS}(X_1), ..., \mathbf{SS}(X_n), \mathbf{SS}(T, X_1), ..., \mathbf{SS}(T, X_n)\}$

2. For each record $R$ in the data set
   (a) Let $t$ denote the value of $T$ in record $R$. (If $t$ does not exist in the linked-list, then $t = t_d$)
   (b) If $t \neq t_d$, increment the one-way count $SS(T = t)$ by one
   (c) For each $(X_i, x_i)$ pair in the linked-list representation for the case, increment the one-way count $SS(X_i = x_i)$ by one. If $t \neq t_d$, increment the two-way count $SS(T = t, X_i = x_i)$ by one.

3. Derive all counts corresponding to default values, using the known number of cases $m$, as follows:
   (a) Derive the one-way counts for each variable $X_i$ (and $T$) using the following equality:

   $$SS(X_i = d_i) = m - \sum_{x_i \neq d_i} SS(X_i = x_i) \quad (1)$$

   (b) Using the one-way counts, derive all two-way counts. In particular, for $x_i \neq d_i$

   $$SS(T = t_d, X_i = x_i) = \quad (2)$$
   $$SS(X_i = x_i) - \sum_{t \neq t_d} SS(T = t, X_i = x_i)$$

   for $t \neq t_d$

   $$SS(T = t, X_i = d_i) = \quad (3)$$
   $$SS(T = t) - \sum_{x_i \neq d_i} SS(T = t, X_i = x_i)$$

   and for $x_i = d_i, t = d_d$

   $$SS(T = t_d, X_i = d_i) = \quad (4)$$
   $$SS(X_i = d_i) - \sum_{t \neq t_d} SS(T = t, X_i = d_i)$$

Figure 3: Algorithm to efficiently extract one-way and two-counts

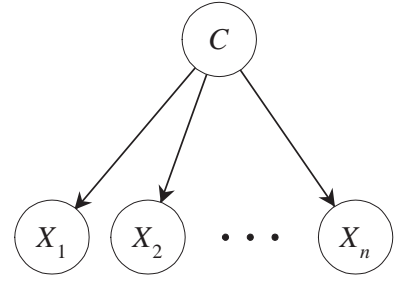

Figure 4: Naive-Bayes model

That is

$$p(C = c|\Theta) = \theta_{C=c}$$

Node $X_i$ stores, for each value $c$ of the cluster variable, a set of $r_i$ parameters $\{\theta_{X_i=1|C=c}, ..., \theta_{X_i=r_i|C=c}\}$ that specify the conditional probability that the variable is in each of its states, given that the record is in cluster $c$. That is,

$$p(X_i = x_i|C = c, \Theta) = \theta_{X_i=x_i|C=c}$$

The EM algorithm repeatedly applies two steps—the E-step and the M-step—until some convergence criterion is met. In the E-step, we use the parameters of the model and the given data to fill in *expected* values for (1) the one-way counts $\mathbf{SS}(C)$ and (2) the two-way counts $\{\mathbf{SS}(C, X_1), ..., \mathbf{SS}(C, X_n)\}$. These resulting expected counts are the expected sufficient statistics for the naive-Bayes model. In the M-step of the EM algorithm, we use the expected sufficient statistics derived in the E-step to set new parameter values $\Theta$ for the model. Dempster et al. (1977) show that the algorithm is guaranteed to converge to a local maximum.

To simplify discussion, we use $x_i^j$ to denote the value of variable $X_i$ in the $j$th record.

In a traditional (dense view) implementation, the expected counts can be calculated as follows. First we initialize all one-way and two-way (expected) counts to zero. Then, for each record $j$ in the data set, we use the parameters of the model to derive the posterior probability of each state of the cluster node:

$$\begin{aligned} p_c^j &= p(C = c|X_1 = x_1^j, ..., X_n = x_n^j, \Theta) \\ &= \alpha \cdot \theta_{C=c} \prod_{i=1}^n \theta_{X_i=x_i^j|C=c} \end{aligned} \quad (5)$$

where $\alpha$ is the normalizing constant. Then, for each value $c$ of the cluster variable, we increment $SS(C = c)$

by $p_c^j$. Similarly, for each value $c$ of the cluster variable, and for each variable $X_i$ in **X**, we increment $SS(C = c, X_i = x_i^j)$ by $p_c^j$. After iterating through all the cases, the one-way and two-way counts contain the expected sufficient statistics for the naive-Bayes model.

We update the parameters of the model using the expected sufficient statistics. For example, if we're identifying the maximum likelihood configuration of the parameters, the M-step will update all of the parameters as follows:

$$\theta_{C=c} = \frac{SS(C=c)}{m}$$

and

$$\theta_{X_i=x_i|C=c} = \frac{SS(X_i = x_i, C = c)}{SS(C = c)}$$

If we're identifying the maximum a posteriori parameter values, a prior term is added to the numerator and denominator in the above equations[1].

In a dense-view implementation of the E-step, we evaluate Equation 5 ($n+1$ multiplications) once for each record, and the resulting algorithm will run in time proportional to $m \cdot n$. (Note that the time to complete the M-step of the EM algorithm does not depend on $m$).

We now describe how to implement the E-step of the EM algorithm in time proportional to the number of values $l$ that are explicitly stored in the sparse representation.

The first speedup, used before each scan of the data set, is to pre-compute the (non-normalized) posterior for each state of the cluster variable, given a (possibly hypothetical) record in which each variable is in the default state. In particular, for each state $c$ of the cluster variable, we compute

$$\begin{aligned} p_c^{default} &= p(C = c, X_1 = d_1, ..., X_n = d_n, \Theta) \\ &= \theta_{C=c} \prod_{i=1}^{n} \theta_{X_i=d_i|C=c} \end{aligned} \quad (6)$$

Given $p_c^{default}$, we efficiently compute $p_c^j$ for each record in the data set by performing only as many multiplications as there are non-default values in the record. The idea is that, for each value in the linked list of a record, we adjust $p_c^{default}$ by multiplying by a corresponding *correction* term $L(x_i, c)$:

---
[1] See (e.g.) Chickering and Heckerman (1997).

$$L(x_i, c) = \frac{\theta_{X_i=x_i|C=c}}{\theta_{X_i=d_i|C=c}} \quad (7)$$

All correction terms $L(x_i, c)$ are pre-computed before each scan of the data. Substituting Equation 7 and Equation 6 into Equation 5 we have

$$p_c^j = \alpha \cdot p_c^{default} \cdot \prod_{x_i \in \text{Record} j} L(x_i, c) \quad (8)$$

where the product is taken over the variables in the record that do not occur in their default values. Note that we still need to evaluate the normalization constant for each record.

The second speedup to the E-step is to apply the methods from Section 3 to efficiently update the expected sufficient statistics of the model. In particular, for each record in the data set, we only update **SS**$(C, X_i)$ for the variables $X_i$ that explicitly appear in the linked list. Note that the one-way counts **SS**$(C)$ are always updated, and that there is no default state for variable $C$. After completing the scan of the data set, we derive the two-way counts for the default values $d_i$ using Equation 3.

Combining these two speedups, we complete the scan of the data in the E-step in time proportional to $l$. Although we incur overhead when (1) pre-computing $p_c^{default}$ and the correction terms and (2) post-computing the expected sufficient statistics corresponding to default values of $X_i$, the time to complete these computations does not depend on the number of records $m$ in the data set. As we see in the following section, the implementation yields dramatic runtime improvements over a dense-view implementation when data is sparse.

## 5 EXPERIMENTS

Our experiments were performed on three real-world data sets.

The first data set, *Television*, uses the Nielsen network television viewing data for a two week period in the winter of 1995. Each variable in the domain corresponds to one of 203 television shows, and each of the 3275 records stores, for each show, whether or not a particular person watched that show. The default value for each show is "didn't watch".

The second data set, *MS Web*[2], contains data on the browsing patterns of people who visited the Microsoft corporate web site in October of 1996. Each of the

---
[2] The MS Web data set is available on the UC Irvine repository.

282 variables in the domain corresponds to a particular URL cluster (vroot) on the site, and each of the 10000 records stores, for each vroot, whether or not a particular person visited that vroot. The default value for each vroot is "didn't visit".

The final data set, *MSNBC*, encodes the stories that people read on the MSNBC web site. The variables correspond to the most popular 2922 stories. The records correspond to the 1,043,878 visitors to MSNBC on one day in December of 1998. Each record stores, for each story, whether or not a particular visitor read that story. The default value for each story is "didn't read".

Table 1 shows summary statistics for each of the data sets. Recall that $n$ is the number of variables, $m$ is the number of records, and $l$ is the number of non-default entries. The last column in the table shows the ratio of the number of logical entries in a dense view of the data ($m \cdot n$) to the number of explicit values stored in a sparse representation of the data ($l$). We expect the running times of the modified algorithms to be faster by an amount that is the same order of magnitude as this ratio.

| Data Set | $n$ | $m$ | $l$ | $\lceil \frac{n \cdot m}{l} \rceil$ |
|---|---|---|---|---|
| Television | 203 | 3,275 | 27,957 | 24 |
| MS Web | 282 | 10,000 | 57,086 | 49 |
| MSNBC | 2,922 | 1,043,878 | 12,348,863 | 247 |

Table 1: Summary statistics for the data sets used in the experiments

Table 2 shows, for each data set, the total time spent executing the E-step of the EM algorithm[3] for two implementations of a naive-Bayes clustering algorithm. The implementations are identical except that one uses a dense view of the data, and the other uses the sparse-data techniques described in Section 4 to optimize the E-step. For each data set, we used a model containing 20 clusters and ran EM to convergence. Because the dense-view implementation was so slow, we relaxed the convergence criterion for the MSNBC data set so that convergence was reached after two iterations of the EM algorithm. The average times reported are from ten separate runs using the same parameters, except for the MSNBC/dense-view experiment which is an average of four runs. The variance in times within each experiment was very small.

Table 3 shows results from applying two implementations of a decision-tree learning algorithm to the data sets; the first implementation uses a dense view of the data, and the second implementation extracts the two-way counts as described in Section 3. The table shows the average total time each algorithm spent extracting two-way counts when scoring splits, where the average was taken over ten separate runs. For each data set, except for MSNBC, we used the algorithms to construct a decision tree for each variable in the domain based on the values for all other variables in the domain. For the MSNBC data set, we restricted the experiment to learn a decision tree for a single variable. The algorithms applied a greedy search strategy and used the Bayes-factor criterion (Chickering et al., 1997).

---
[3]We report only the cumulative times spent in the E-step, but for large data sets such as MSNBC, this time dominates the total running time of the EM algorithm

| Data Set | Dense (ms) | Sparse (ms) | $\lceil$ Ratio $\rceil$ |
|---|---|---|---|
| Television | 76,776 | 5,454 | 14 |
| MS Web | 397,152 | 9,901 | 40 |
| MSNBC | 28,301,773 | 519,250 | 55 |

Table 2: Clustering results

| Data Set | Dense (ms) | Sparse (ms) | $\lceil$ Ratio $\rceil$ |
|---|---|---|---|
| Television | 44848 | 3342 | 13 |
| MS Web | 368005 | 10613 | 35 |
| MSNBC | 1674595 | 13188 | 127 |

Table 3: Decision-tree results

As expected, sparser data (i.e. larger values of $\frac{m \cdot n}{l}$) yielded larger improvements in running time for the modified algorithms. Furthermore, the relative speedups in Tables 2 and 3 are of the same order of magnitude as the corresponding ratios $\frac{n \cdot m}{l}$ shown in Table 1. The most dramatic improvement was achieved when learning decision trees for the MSNBC dataset: our sparse implementation of the algorithm took roughly 13 seconds to learn a decision tree, whereas the dense-view implementation took almost a half hour.

## 6 CONCLUSION

In this paper, we described two methods for speeding up several machine-learning algorithms when data is sparse. We demonstrated that these methods yield dramatic improvements in running time for both clustering algorithms and decision-tree learning algorithms when applied to three real-world data sets. The methods are generally applicable to algorithms that extract one-way and two-way counts from the data.

# References

Breiman, L., Friedman, J., Olshen, R., and Stone, C. (1984). *Classification and Regression Trees.* Wadsworth & Brooks, Monterey, CA.

Buntine, W. (1991). Theory refinement on Bayesian networks. In *Proceedings of Seventh Conference on Uncertainty in Artificial Intelligence,* Los Angeles, CA, pages 52–60. Morgan Kaufmann.

Cheeseman, P. and Stutz, J. (1995). Bayesian classification (AutoClass): Theory and results. In Fayyad, U., Piatesky-Shapiro, G., Smyth, P., and Uthurusamy, R., editors, *Advances in Knowledge Discovery and Data Mining*, pages 153–180. AAAI Press, Menlo Park, CA.

Chickering, D. and Heckerman, D. (1997). Efficient approximations for the marginal likelihood of Bayesian networks with hidden variables. *Machine Learning*, 29:181–212.

Chickering, D., Heckerman, D., and Meek, C. (1997). A Bayesian approach to learning Bayesian networks with local structure. In *Proceedings of Thirteenth Conference on Uncertainty in Artificial Intelligence,* Providence, RI. Morgan Kaufmann.

Clogg, C. (1995). Latent class models. In *Handbook of statistical modeling for the social and behavioral sciences*, pages 311–359. Plenum Press, New York.

Cooper, G. and Herskovits, E. (1991). A Bayesian method for constructing Bayesian belief networks from databases. In *Proceedings of Seventh Conference on Uncertainty in Artificial Intelligence,* Los Angeles, CA, pages 86–94. Morgan Kaufmann.

Dempster, A., Laird, N., and Rubin, D. (1977). Maximum likelihood from incomplete data via the EM algorithm. *Journal of the Royal Statistical Society*, B 39:1–38.

Edmonds, J. (1967). Optimum branching. *J. Res. NBS*, 71B:233–240.

Heckerman, D., Geiger, D., and Chickering, D. (1995). Learning discrete Bayesian networks. *Machine Learning*, 20:197–243.

Moore, A. and Lee, M. S. (1998). Cached sufficient statistics for efficient machine learning with large datasets. *Journal of Artificial Intelligence Research*, 8:67–91.

Spiegelhalter, D., Dawid, A., Lauritzen, S., and Cowell, R. (1993). Bayesian analysis in expert systems. *Statistical Science*, 8:219–282.